\documentclass[lettersize,journal]{IEEEtran}

\expandafter\def\expandafter\normalsize\expandafter{%
    \normalsize
    \setlength\abovedisplayskip{4pt plus 2pt minus 2pt}
    \setlength\belowdisplayskip{4pt plus 2pt minus 2pt}
    \setlength\abovedisplayshortskip{2pt plus 1pt minus 1pt}
    \setlength\belowdisplayshortskip{2pt plus 1pt minus 1pt}
}

\usepackage{amsmath,amsfonts}
\usepackage{algorithmic}
\usepackage{algorithm}
\usepackage{array}
\usepackage[caption=false,font=normalsize,labelfont=sf,textfont=sf]{subfig}
\usepackage{textcomp}
\usepackage{stfloats}
\usepackage{url}
\usepackage{verbatim}
\usepackage{graphicx}
\usepackage{booktabs} 
\usepackage{xcolor}
\usepackage{multirow}

\usepackage[capitalise]{cleveref} 

\begin{document}

\title{Belief-Aware Influence and Trust (BAIT): Shaping Human Belief During Repeated Human-Robot Interaction}

\author{Ye-Ji Mun$^1$, Mahsa Golchoubian$^2$, Shahabedin Sagheb$^3$, Yan Bai$^1$, Tianhao Ji$^1$, \\  Dylan P. Losey$^3$, and Katherine Driggs-Campbell$^1$
\thanks{$^{1}$ University of Illinois at Urbana-Champaign, Department of Electrical and Computer Engineering.}
\thanks{$^{2}$ University of Toronto, Department of Mathematical and Computational Science.}
\thanks{$^3$ Virginia Tech, Department of Mechanical Engineering.}
\thanks{Corresponding author's email: \texttt{yejimun2@illinois.edu}}
}



\maketitle

\begin{abstract}
Repeated human-robot interaction (HRI) requires proactively accounting for humans who continually adapt to evolving beliefs about the robot. Prior frameworks often treat encounters as isolated events, suffering cumulative task performance decay as human perception drifts, or maintain long-term influence through erratic, unpredictable behavior that erodes perceived human trust and relies on computationally unscalable formulations. To address these gaps, we introduce the Belief-Aware Influence and Trust (BAIT) controller. BAIT integrates a hierarchical particle filter, which infers both fast human strategic shifts and slow perceptual belief updates, with a belief-aware Model Predictive Path Integral planner. BAIT explicitly optimizes the trade-off between long-horizon influence and human trust, while enforcing immediate task performance as a strict constraint. Across simulations, a human-subject study, and a real-world GEM vehicle deployments in repeated lane-merging scenarios, BAIT achieves task performance comparable to baselines that optimize long-term influence through unpredictability while yielding significantly higher user trust. The video demonstrating our experiments is available at https://youtu.be/9o4GqKLWDCw.

\end{abstract}


\section{Introduction}

As robots become integrated into human spaces, they must account for an unavoidable dynamic: human behavior continually adapts in response to robot actions, especially across repeated encounters. Consider a delivery robot crossing paths with the same office workers every morning, an autonomous vehicle merging past the same commuters, or an assistive robot navigating around a household. In each of these settings, human behavior is neither static nor purely repetitive but rather evolves as humans observe the robot. When robots are predictably defensive, humans often exploit this predictability (e.g., human drivers refusing to let a hesitant vehicle merge or pedestrians asserting right-of-way), degrading task efficiency and efficacy. Sustained task performance across repeated encounters therefore depends on how a robot's immediate actions shape the human's evolving perception and future behavior.

\begin{figure}[t] 
    \centering
    \includegraphics[width=\linewidth]{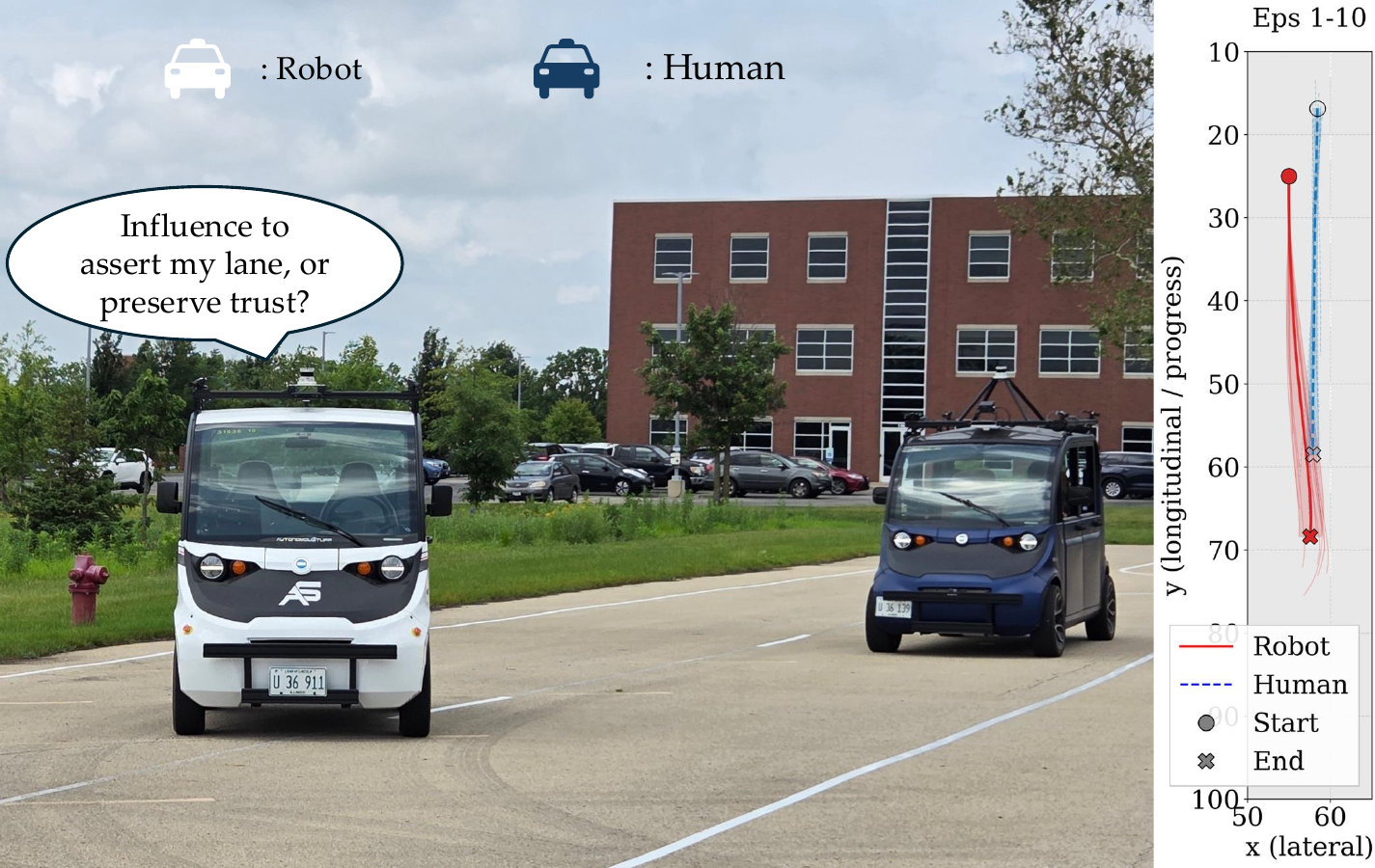}
    \vspace{-0.6cm}
    \caption{\textbf{Repeated lane-merging encounters deployed on Polaris GEM vehicles.} Our BAIT-adapt robot (white) explicitly navigates the internal trade-off between influence to assert its lane right-of-way and preserving human trust. As shown in the corresponding trajectory rollouts (right), the BAIT controller (solid \textcolor{red}{red}) successfully executes the merge across repeated $10$ episodes while consistently prompting the human driver (dashed \textcolor{blue}{blue}) to yield.} 
    \vspace{-0.5cm}
    \label{fig:GEM}
\end{figure}
\IEEEpubidadjcol

Existing human-aware robot planning frameworks frequently overlook this human dynamic, treating each encounter as an isolated event and modeling the human as a fixed reward-maximizing agent~\cite{mun2023pas}, a static behavior distribution~\cite{Xie2021latentinfluence, Sun2025mixednash}, or an offline-trained trajectory predictor~\cite{Huang2022trajpredsurvey}. These stationary assumptions may be sufficient for one-shot interactions but fundamentally inadequate in the long-term settings described above. A planner that ignores how its current robot behavior alters future human behaviors optimizes only the immediate interaction, and empirically suffers from measurable task-performance decay as the human's perception drifts. 

A few recent works instead model the evolution of human internal state to actively guide this dynamics to maintain cumulative task performance~\cite{sagheb2025unified, sagheb2023towards}, but face two primary limitations that constrain their deployment as practical controllers. First, these approaches maintain long-horizon influence by deliberately generating unpredictable robot behavior. Such unpredictability provides a reasonable initial mechanism for preventing humans from settling into a lazy or low-effort response pattern, thereby preserving the robot's influence over repeated interactions. However, this mechanism comes at a cost: the resulting robot behavior may be erratic, which degrades behavioral transparency and reduces human trust~\cite{bastarache2023legible,dragan2013legibility}. We show that the unpredictability of previous long-term influence approaches narrows safety margins and erodes perceived human trust and comfort. Second, these long-term influence controllers face severe scalability bottlenecks relying on either strict leader-follower Stackelberg assumptions~\cite{sagheb2023towards} or computationally heavy Mixed Observability Markov Decision Process (MOMDP) solvers~\cite{sagheb2025unified}.

To address these scalability and unpredictability limitations, we make the following contributions:

\begin{itemize}
    \item \textbf{The BAIT framework.} We propose the \textbf{Belief-Aware Influence and Trust (BAIT)} controller, which explicitly trades off long-horizon influence (unpredictability) against human trust (transparency). BAIT structures the human's latent state into a two-timescale hierarchy: a short-term strategy variable ($z^t$) to predict immediate human behaviors and a long-term perception variable ($\phi^t$) to capture the human's evolving beliefs of the robot.
 
    \item \textbf{Online Influence-Trust Management.} We track the human's evolving belief in real time using a two-layered particle filter that fuses action and geometric evidence. This inferred latent state is propagated forward as a deterministic surrogate into a belief-aware Model Predictive Path Integral (MPPI) planner. This integration enables BAIT to uniquely optimize the influence-trust trade-off as a primary design axis, rather than optimizing either objective in isolation, while enforcing immediate task performance as a hard constraint.

    \item \textbf{Sustained Influence and Improved Human Trust.} Through simulation and real-world hardware experiments, we demonstrate that BAIT sustains higher cumulative task performance than influence-only baselines. A user study further shows that BAIT significantly improves human trust over influence-only controllers while preserving the task performance that trust-only methods, which prioritize transparent robot behavior, sacrifice.
\end{itemize}
\section{Related Work} \label{sec:related_work}

\subsection{Influence in HRI: Short-Term to Long-Term}
Traditional frameworks for HRI often treat encounters as an interdependent game where the robot acts proactively to shape human behavior~\cite{sadigh2016planning,fisac2019hierarchical,schwarting2019social,parekh2023learning}. These proactive influence approaches are primarily optimized for single or short-term interactions, frequently employing Stackelberg formulation to maximize immediate task efficiency. However, these static models suffer from performance decay during repeated interactions because humans continually adapt over time, eventually exploiting or ignoring predictable robot behaviors \cite{cooper2019stackelberg, ayub2025continual}. There have been recent attempts to model this long-term adaptation. Sagheb et al.~\cite{sagheb2023towards} address long-horizon influence by deliberately generating unpredictable robot behaviors to prevent humans from exploiting predictable robot policies rather than coordinating with the robot. However, their method relies on the Stackelberg formulation, which is based on best-response human and leader-follower assumptions that are not realistic in the real world. Sagheb et al.~\cite{sagheb2025unified} propose a framework that unifies these repeated interactions through a MOMDP formulation and show how prior methods are simplifications of this unified framework. However, these approaches rely on computationally intensive solvers. Standard online MOMDP solvers (e.g., POMCPOW) evaluate full belief trees, creating computational bottlenecks as state and action dimensions scale. We extend this lineage by maintaining long-horizon influence while ensuring real-time computational tractability via a conditionally decoupled hierarchical formulation. 

\subsection{Transparent Actions and Human Trust}
Foundational human factors research establishes that legible and predictable robot trajectories are crucial for fostering user trust and comfort~\cite{dragan2013legibility, bastarache2023legible}. However, this requirement creates a fundamental tension with recent proactive influence methods, which deliberately generate unpredictable robot behavior to force human compliance \cite{sagheb2025unified, sagheb2023towards} over repeated interactions. To mitigate this erosion of trust, Peng et al.~\cite{peng2025trust} employ Bayesian Persuasion to derive a theoretical minimum trust level required to influence humans in a transparent way. However, this approach optimizes for a static human model, which may not fully capture scenarios where human perception continuously evolves.
In this work, we address this influence-trust tension for dynamic human adaptation by explicitly treating the trade-off between unpredictability-driven influence and trust as our primary design axis, maintaining the robot's necessary influence while preserving human trust.

\subsection{Human Modeling with Latent Representations}
To capture the hidden intent of human partners, a prominent line of work represents their strategies using continuous or discrete latent variables~\cite{xie2021learning, bajcsy2024learning}. These methods infer continuous or discrete latent embeddings online to adapt robot policies within an interaction. However, most existing latent representation frameworks operate under the assumption that the underlying dynamics of human adaptation are stationary, often reducing the problem to a one-step Q-function for Markov Decision Process that ignores exploratory information gathering~\cite{sagheb2025unified}. In this work, we structure the human's hidden state into a conditionally coupled, two-timescale hierarchy that separates high-frequency tactical reactions from gradual perceptual drift. This hierarchical separation bypasses the inefficiency of standard flat filters and allows the robot to actively plan over the human's evolving belief~\cite{huang2025hierarchical}.

\section{Problem Statement} \label{sec:ProblemStatement}
We study robot decision-making in repeated human-robot interactions, focusing specifically on interactive lane changing in autonomous driving. When vehicles repeatedly negotiate shared lane space and right-of-way, humans do not act passively; they continuously observe the robot, update their internal perception of the robot, and adapt their strategy accordingly. Unlike one-shot encounters where human policies can be approximated as stationary~\cite{ratliff2018perspective, foerster2018learning}, repeated interactions induce non-stationary behavior across distinct timescales as the human adapts with their beliefs about the robot's behavior. Consequently, this adaptation requires a dual objective for the robot: (i) accurately inferring the underlying latent states, such as human's belief and behavioral strategy, and (ii) leveraging the inference for belief-aware planning. Specifically, the planner must steer the evolution of the human's perception toward configurations that secure favorable long-term outcomes without violating immediate task constraints. Crucially, in traffic navigation, securing these outcomes means safeguarding the robot's  influence to prevent behavioral exploitation, ensuring the robot retains the leverage necessary to induce human cooperation, such as yielding during a lane change.

To capture human adaptation, we model the human's internal state using a hierarchy of two tightly coupled latent variables: a fast-changing short-term strategy $z^t$ and a slow-evolving long-term belief $\phi^t$. The \emph{short-term strategy} $z^t{\in}\mathbb{R}^d$ encodes the human's instantaneous weighting vector across $d$ competing task-related costs, such as forward progress ($z_{\text{progress}}$) versus collision avoidance ($z_{\text{safety}}$), enabling rapid reactions to the immediate physical context. Conversely, the \emph{long-term belief} $\phi^t {\in} [0,1]$ encodes the human's overarching perception of the robot, ranging from assertive ($\phi{\rightarrow}0$) to defensive ($\phi{\rightarrow}1$), and evolves gradually based on accumulated evidence across multiple interactions. Crucially, $\phi^t$ acts as a prior that modulates $z^t$ formally captured by the conditional probability $p(z^t{\mid}\phi^t)$. For example, during a lane merge, perceiving the robot as defensive ($\phi{\rightarrow}1$) induces an aggressive momentary strategy ($z_{\text{progress}}{>}z_{\text{safety}}$), whereas perceiving an aggressive robot ($\phi\rightarrow0$) prompts conservative yielding ($z_{\text{safety}} {>} z_{\text{progress}}$) to prioritize safety.

Building upon the Unified Framework for long-term influence proposed by~\cite{sagheb2025unified}, we formulate the problem as a MOMDP governed by the two-timescale latent hierarchy. However, unlike prior work that optimizes the MOMDP solely for cumulative influence, we introduce a scalable, real-time architecture that explicitly balances influence against human trust. Let $s^t{=}\left[s_\mathcal{R}^t, s_\mathcal{H}^t\right]$ denote the fully observable physical state of the robot $\mathcal{R}$ and the human $\mathcal{H}$. The unified MOMDP state is defined as $x^t {=} \left[ s^t, z^t, \phi^t \right]$ with transitions defined by the joint dynamics:
\begin{align}\label{eq:momdp}
    x^{t+1} &= \begin{bmatrix} s^{t+1} \\ z^{t+1} \\ \phi^{t+1} \end{bmatrix} = 
    \begin{bmatrix}  
    f_p\big(s^t, a^t \big)
    \\ f_s\big(s^t, a^t, z^t, \phi^t \big)
    \\ f_l\big(s^t, a^t, \phi^t \big) 
    \end{bmatrix} 
    = F (x^t, a_\mathcal{R}^t)
\end{align}
where $a^t {=} \left[a_\mathcal{R}^t, a_\mathcal{H}^t \right]$ represents the joint robot and human action, $f_p$ defines the physical dynamics, $f_s$ is the short-term strategy transition, and $f_l$ is the long-term, belief transition. Substituting the human policy into equation~\ref{eq:momdp} yields the underactuated dynamics $F (x^t, a_\mathcal{R}^t)$ from the robot's perspective, where the human action $a_\mathcal{H}^t {\sim} \pi_\mathcal{H}(\cdot {\mid} s^t, z^t)$ is drawn from a maximum-entropy Boltzmann policy:
\begin{equation} \label{eq:boltzmann_policy}
    \pi_\mathcal{H}(a \mid s, z) \propto\exp \big( -\eta \cdot z^\top \bar r(s,a_{\mathcal{R}}, a_{\mathcal{H}}) \big)  
\end{equation}
where $\bar r {\in} \mathbb{R}^d$ is a vector of instantaneous physical reward features parameterized linearly by the strategy weights $z$, and $\eta$ is a rationality parameter where $\eta {\rightarrow} \infty$ recovers full rational behavior. A crucial structural property of these dynamics is that $\phi^t$ only influences actions indirectly through $z^t$, creating a partial-evidence problem where the human's true perception is masked by their immediate strategy.

\begin{figure*}[ht!]
    \centering
    \vspace{-0.5cm}  
    \includegraphics[width=\linewidth]{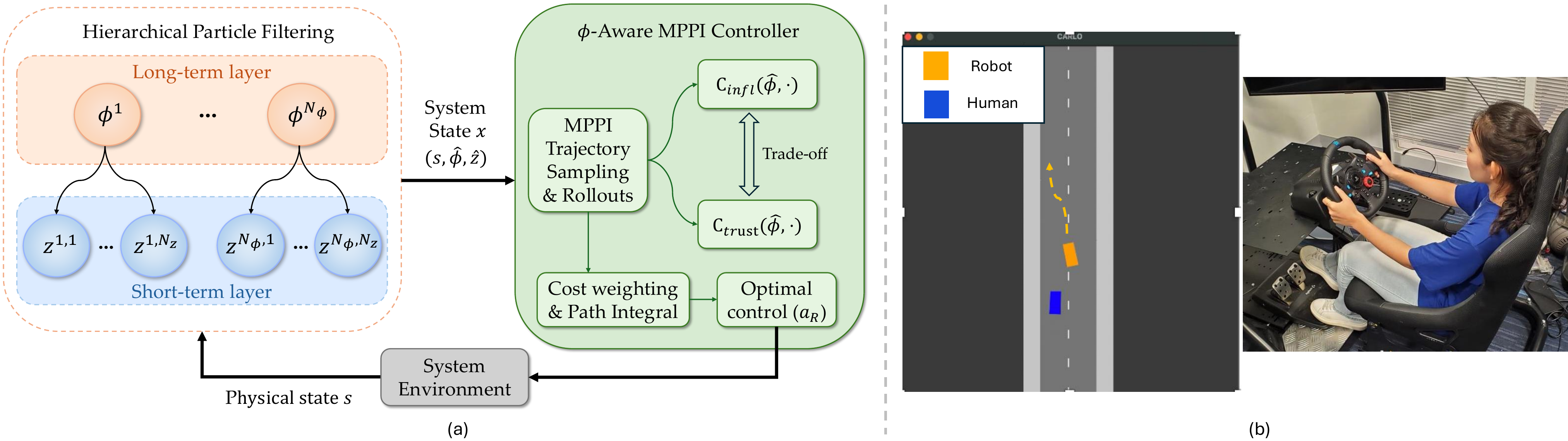}
    \vspace{-0.5cm}    
    \caption{\textbf{Overview of our BAIT controller.} (a) Our BAIT controller has two tightly coupled modules: a hierarchical particle filter that tracks the joint posterior over $(\phi,z)$ from observed interaction history, and an MPPI planner that rolls this posterior forward through a deterministic belief-propagation using the posterior mean $(\hat \phi,\hat z)$. This planner optimizes an explicit influence-trust trade-off while enforcing a hard task-cost constraint. (b) The 2D CARLO lane-merging simulation environment. }
    \label{fig:bait_overview}
    \vspace{-0.5cm}    
\end{figure*}

The robot's objective is to solve this MOMDP over a receding planning horizon $H$ by actively shaping $\phi^t$ to maximize cumulative task performance. Formally, the robot seeks an action sequence $\mathbf{a}_\mathcal{R} {=} a_\mathcal{R}^{t:t+H}$ that optimizes a joint objective:
\begin{align} \label{eq:general_objective}
    \min_{\mathbf{a}_\mathcal{R}} \quad & \mathbb{E} \left[ \sum_{k=t}^{t+H} \mathcal{C}_{\text{task}}(s^k, a_\mathcal{R}^k) + \mathcal{C}_{\text{belief}}(\phi^{k+1}, s^k, a_\mathcal{R}^k) \right] \\
    \text{s.t.} \quad & x^{k+1} = F(x^k, a_\mathcal{R}^k) \nonumber
\end{align}
To maintain influence while preserving human trust, the latent cost $\mathcal{C}_{\text{belief}}$ explicitly balances robot's behavioral unpredictability against transparency. Specifically, $\mathcal{C}_{\text{belief}}$ penalizes states where the human perceives the robot as easily exploitable ($\phi {\rightarrow} 1$), such as when the human perceives the robot will act passively defensive. Simultaneously, this cost regularizes against erratic or overly unpredictable actions, ensuring that the robot's overarching strategy remains sufficiently transparent to preserve human trust during the traffic negotiation.

\section{Belief-Aware Influence and Trust} \label{sec:method}

To solve the Unified MOMDP formalism in~\cref{sec:ProblemStatement}, we introduce the Belief-Aware Influence and Trust (BAIT) controller. Whereas prior long-term influence approaches focus on maximizing cumulative influence, often generating unpredictable behaviors that degrade human trust, BAIT explicitly optimizes the trade-off between proactive influence and trust (operationalized through behavioral transparency). To actively evaluate this dual objective in real time, BAIT overcomes the computational bottlenecks inherent to continuous-space MOMDPs by decoupling the problem into two tightly integrated modules: a hierarchical particle filter that tracks the joint posterior over $(\phi^t,z^t)$ from observed interactions, and an MPPI planner that rolls this inferred posterior forward using deterministic belief-propagation as shown in~\cref{fig:bait_overview}.

\subsection{Hierarchical Inference of Human Latent States}  \label{sec:particle_filter}
We track the human's hidden state using a two-layered particle filter designed to preserve the timescale separation and conditional structure $p(z|\phi)$ of the latent hierarchy formalized in~\cref{sec:ProblemStatement}. While a standard flat filter samples $(\phi, z)$ jointly, such a uniform structure scales inefficiently by enforcing redundant belief evaluations during high-frequency strategy $z$ updates and causing a 
computational bottleneck. To resolve this, we structure the posterior as a hierarchical ensemble:
\begin{equation} \label{eq:pf_state}
\mathcal{B}^t = \big\{ \phi^{(i)}, w_\phi^{(i)}, \{z^{(i,j)}, w^{(i,j)}_z
\}^{N_z}_{j=1}  \big\}^{N_\phi}_{i=1}
\end{equation}
where each  of the $N_\phi$ long-term particles carries its own conditional ensemble of $N_z$ short-term particles.

\noindent\textit{
\textbf{Prediction Step:}} The prediction step reflects this hierarchy by applying distinct transition kernels matched to each variable's timescale. The long-term belief variable $\phi$ propagates via a mixture Beta kernel encoding bounded, gradual drift alongside a slow reset probability $\epsilon$ to prevent particle collapse:
\begin{equation}
    \phi^{t+1} \sim (1-\epsilon) \text{Beta} \big( \kappa \phi^t, \kappa (1-\phi^t) \big) + \epsilon \text{Beta}(c,c)
\end{equation}
where $\kappa$ is a concentration parameter that controls the magnitude of the step-to-step drift variance (i.e., a higher $\kappa$ yields a tighter concentration and lower variance), while $c$ shapes the distribution of the random reset particles (e.g., $c{=}1$ for a uniform reset). Conditioned on this predicted belief $\phi^{t+1}$, the short-term strategy $z$ evolves via an Ornstein-Uhlenbeck (OU) process that mean-reverts toward an attractor $z_{\text{tgt}}$ determined by its parent $\phi^{t+1}$:
\begin{equation} \label{eq:z_ou}
    z^{t+1} = z^t + \alpha_{\text{ou}} \big( z_{\text{tgt}} (\phi^{t+1})-z^t \big) +\sigma_z \xi^t, \quad \xi^t \sim \mathcal{N}(0,I)
\end{equation}
where $\alpha_{\text{ou}} {\in} (0,1]$ is the mean-revision rate that dictates how rapidly the human's short-term strategy adapts toward the target attractor $z_{tgt}(\phi^{t+1})$, and $\sigma_z$ scales the standard normal variable $\xi^t$ to model the stochasticity and natural execution variance inherent to human driving behavior.
This coupling structurally enforces the behavioral prior established in Section~\ref{sec:ProblemStatement} (e.g., drawing $z$ towards an aggressive momentary strategy when $\phi{\rightarrow}1$ and conservative strategy $\phi{\rightarrow}0$).

\noindent\textit{
\textbf{Update Step:}} To robustly update these particles despite the indirect observability, the filter's update step applies distinct evidence to each hierarchical layer. At the bottom layer, the short-term $z$-particles are reweighted using \emph{action evidence} via the Boltzmann log-likelihood of the observed human actions in~\cref{eq:boltzmann_policy} over a decaying history length $W$:
\begin{equation}\label{eq:z_logll_history}
\ell_z^{(i,j)} \;=\; \sum_{\tau=t-W+1}^{t} \rho^{\,t-\tau}\,\log \pi_\mathcal{H}\!\big(a^\tau_\mathcal{H} \,\big|\, s^\tau,\, z^{(i,j)}\big)
\end{equation}
where $\rho {\in} (0,1]$ is an exponential decay factor. To prevent premature weight collapse under low-signal observations, we update the normalized weight using a soft inertia blend:
\begin{equation} \label{eq:z_weight_update}
w_z^{(i,j)} \leftarrow (1-\alpha_z) w_z^{(i,j)} + \alpha_z \frac{\exp(\ell_z^{(i,j)})}{\sum_{m=1}^{N_z} \exp(\ell_z^{(i,m)})}
\end{equation}
where $\alpha_z$ is the update rate for the new evidence. 

At the top layer, the long-term $\phi$-particles are updated by fusing this indirect \emph{action evidence} with a direct \emph{geometric evidence} channel (e.g., lateral gap and closing rate for lane merging). First, the action evidence is aggregated for each parent $\phi^{(i)}$ by marginalizing the log-likelihood of its $z$-ensemble in Eq.~\ref{eq:z_logll_history}: 
\begin{equation} \label{eq:phi_logll_history}
\ell_\phi^{(i)} = \log \sum_{j=1}^{N_z} \exp\big(\ell^{(i,j)}\big).
\end{equation}
Simultaneously, a geometric proxy infers the human's belief based on recent physical interactions; for example, if the robot merges aggressively, it naturally induces the human to believe the robot will not yield. We extract an instantaneous yield-belief target $y^t$ from these kinematic features and maintain a smooth yield-belief $b_y^t$ via an exponential moving average: 
\begin{equation}\label{eq:belief_proxy}
b_y^t \leftarrow (1-\alpha_\phi)b_y^{t-1} + \alpha_\phi y^t
\end{equation}
where $\alpha_\phi$ is the smoothing rate. 
The filter fuses both evidence channels in log-space to update the intermediate $\phi^t$ weights: 
\begin{equation} \label{eq:phi_weight_update}
\log \tilde{w}_\phi^{(i)} = \log w_\phi^{(i)} + \frac{1}{T_\phi} \ell_\phi^{(i)} + \frac{\mathbb{I}_{\text{int}}}{T_\phi} \log p\big(b_y^t \mid \phi^{(i)}\big)
\end{equation}
where $T_\phi$ is a scaling temperature, $p(b_y^t {\mid} \phi^{(i)})$ represents the likelihood of the particle given the geometric evidence, and $\mathbb{I}_{\text{int}}$ is an indicator function that activates the geometric channel only when the human enters a critical interaction zone defined by spatial threshold $d_{\text{interact}}$ and the robot is longitudinally ahead of the human driver. In the context of lane merging, this threshold prevents nominal driving maneuvers from miscalibrating the belief $\phi^t$ by ensuring geometric evidence is only fused when vehicles are close enough to actively negotiate the right-of-way. Normalizing these intermediate weights $\tilde{w}_\phi^{(i)}$ yields the final posterior weights $w_\phi^{(i)}$. This dual-channel fusion yields a joint posterior that remains well-calibrated across varying interaction intensities, effectively resolving the partial-evidence bottleneck. 

\noindent \textit{
\textbf{Resampling Step:}} Finally, resampling is triggered independently for each layer based on an effective-sample-size (ESS) criterion, respecting the timescale hierarchy. The long-term $\phi$-layer employs Liu-West kernel smoothing to maintain stable continuous moments. Conversely, the short-term $z$-ensembles are resampled with a partial injection of new particles drawn from the prior $p(z|\phi^{(i)})$ to ensure the strategy remains agile to rapid behavioral shifts.

\subsection{Belief-Aware MPPI Planning} \label{sec:planning}
Leveraging the inferred posterior ($\hat{\phi}, \hat{z}$), our belief-aware MPPI planner actively optimizes the robot's control sequence by explicitly trading off long-term influence against human trust. At each time step, the controller samples $K$ candidate action sequences and evaluates them against a belief cost, $\mathcal{C}_{\text{belief}}$. This cost dynamically balances two opposing terms using an arbitration parameter $\beta {\in} [0,1]$ that shifts the controller from pure trust ($\beta{=}0$) to pure influence ($\beta {=} 1$):
\begin{equation}\label{eq:BAIT-adapt}
\mathcal{C}_{\text{belief}}  = \big( \beta \mathcal{C}_{\text{infl}} + (1-\beta) \mathcal{C}_{\text{trust}}  \big) \mathbb{I}_{\text{int}}  
\end{equation}
where $\mathbb{I}_{\text{int}}$ is the same spatial interaction indicator defined in~\cref{eq:phi_weight_update}. Through this gating, the robot restricts its belief-shaping maneuvers to active lane-changing negotiations, avoiding unnecessary adjustments in open traffic and acting only when the human driver's internal state actively evolves.


The two belief-aware terms in~\cref{eq:BAIT-adapt} are anchored on propagated entropy of the human belief regarding the robot's yielding intent $\mathcal{H}(\hat{\phi}^{k+1})$. The influence cost $\mathcal{C}_{\text{infl}}$ promotes unpredictable trajectories ($\mathcal{H}\uparrow$) to prevent the human from settling into an assertive policy, thereby inducing conservative yielding. However, optimizing for entropy in isolation could produce erratic, task-irrelevant behavior (e.g., randomly oscillating across lanes). To prevent this, we introduce a directional regularizer. Weighted heavily below the entropy ($w_d {\ll} w_e$), this regularizer biases the unpredictability toward assertive outcome ($\phi {\rightarrow} 0$), ensuring the robot's actions actively claim right-of-way:
\begin{equation} \label{eq:cost_infl}
\mathcal{C}_{\text{infl}} = w_e\big(1-\mathcal{H}(\hat{\phi}^{k+1})\big)^2+w_d(\hat{\phi}^{k+1})^2.
\end{equation}
In contrast, the trust cost $\mathcal{C}_{\text{trust}}$ promotes \emph{transparent} trajectories whose induced perception is predictable ($\mathcal{H}(\hat\phi^{k+1}){\downarrow}$), signaling an accommodating intent ($\phi {\rightarrow} 1$) via a legible merge:
\begin{equation} \label{cost_tran}
\mathcal{C}_{\text{trust}} = w_e\mathcal{H}(\hat{\phi}^{k+1})^2+w_d(1-\hat{\phi}^{k+1})^2.
\end{equation}
Since $\mathcal{H}(\hat{\phi})$ is symmetric, $w_d {\ll} w_e$ only sets belief direction, not transparency (ablated in~\cref{sec:simulation}).
To evaluate $\mathcal{C}_{\text{infl}}$ and $\mathcal{C}_{\text{trust}}$ across $K$ candidate sequences and $H$ horizon steps, running the full sample-based filter is computationally intractable. Instead, we ensure real-time scalability by propagating a noise-free \emph{deterministic surrogate} of the posterior moments. We initialize each rollout with the current posterior means, $\hat\phi^t{=}\mathbb{E}[\phi^t {\mid} \mathcal{B}^t]$ and $\hat z^t{=}\mathbb{E}[z^t {\mid} \mathcal{B}^t]$, alongside the current yield-belief $b^t_y$. At each horizon step $k {\in} [t, t+H-1]$, we propagate these posteriors forward via three operations: (i) predicting the human action using the Boltzmann policy $a^k_{\mathcal{H}} {\sim} \pi_\mathcal{H}(\cdot {\mid} s^k, \hat z^k)$; (ii) advancing physical state through the joint dynamics from \cref{eq:momdp}; and (iii) updating the surrogate state using the geometric kernel from~\cref{eq:belief_proxy} and the noise-free strategy dynamics from~\cref{eq:z_ou}. Our BAIT approach is explicitly reliant on this structured human model, and this efficient forward pass tracks the expected latent state, providing the necessary signals to compute the belief-aware costs without the computational burden of a full ensemble simulation.

Finally, to ensure that belief-shaping maneuvers do not compromise physical safety or primary task objectives, the planner treats the task cost ($\mathcal{C}_{\text{task}}$ in Eq.~\ref{eq:general_objective}), which encompasses collision avoidance and lane merging, as a strict conditional value-at-risk (CVaR) constraint:
\begin{equation}\label{eq:task_constraint}
\mathcal{C}_{\mathrm{task}} = M\,\mathbb{I}\big[\mu^{(k)}_{\text{task}} + \eta_r\,\sigma^{(k)}_{\text{task}} > \tau\big]
\end{equation}
where $M$ is a sufficiently large scalar penalty, $\mu^{(k)}_{\text{task}}$ and $\sigma^{(k)}_{\text{task}}$ denote the mean and standard deviation of the task cost for the $k$-th rollout, $\eta_r$ is a risk-aversion parameter, and $\tau$ is the acceptable cost threshold. Any rollout whose upper-bound risk assessment exceeds this threshold receives the penalty $M$, driving its corresponding MPPI sampling weight to near-zero and functionally excluding the unsafe trajectory from the control update. This architecture ensures that BAIT exercises its influence-trust trade-off strictly within the set of physically safe and task-feasible trajectories.

\section{Simulation Experiment}\label{sec:simulation}
We evaluate BAIT in a repeated lane-merging scenario with simulated humans. The robot starts ahead and merges into the human's lane. If the robot influences the human to yield (low lane progress), the merge succeeds; otherwise, the human maintain or increases speed, blocking the merge. This experiment aims to assess two core algorithm capabilities: (i) the invariant estimation accuracy of the two-layered particle filter against ground-truth human parameters across different robot control modes, and (ii) the impact of the influence--trust arbitration on cumulative task performance against an adapting human agent. While this section strictly assesses objective algorithmic metrics, subjective evaluations of human trust are addressed through a user study in Section~\ref{sec:user_study}.

\begin{table*}[t]
\centering
\caption{Robustness of the Hierarchical Particle Filter Estimation Across Varying Robot Control Modes.}
\label{tab:pf_performance}
\small
\setlength{\tabcolsep}{4pt} 
\begin{tabular}{llccccccc}
\toprule
& & \multicolumn{4}{c}{\textbf{Latent State Estimation}} & \multicolumn{3}{c}{\textbf{Trajectory Prediction}} \\
\cmidrule(lr){3-6} \cmidrule(lr){7-9}
\textbf{Environment} & \textbf{Method} & 
\textbf{$\phi$ BCE} $(\downarrow)$ & 
\textbf{$\phi$ MSE} $(\downarrow)$ & 
\textbf{$z$ MSE} $(\downarrow)$ & 
\textbf{$z$ Cos.} $(\uparrow)$ & 
\textbf{1-ADE} $(\downarrow)$ &
\textbf{H-ADE} $(\downarrow)$ & 
\textbf{H-FDE} $(\downarrow)$ \\
\midrule
\multirow{3}{*}{\begin{tabular}[c]{@{}c@{}}\textit{Simulated} \\ \textit{Human}\end{tabular}}& BAIT-trust & $0.44 (0.25)$ & $0.04 (0.03)$ & $0.09 (0.10)$ & $0.95 (0.04)$ & $0.050 (0.010)$ & $0.530 (0.060)$ & $1.100 (0.140)$ \\
& BAIT-infl & $0.69 (0.25)$ & $0.05 (0.02)$ & $0.04 (0.03)$ & $0.97 (0.02)$ & $0.050 (0.010)$ & $0.510 (0.070)$ & $1.070 (0.140)$ \\
& BAIT-adapt & $0.52 (0.31)$ & $0.05 (0.03)$ & $0.07 (0.09)$ & $0.96 (0.04)$ & $0.050 (0.010)$ & $0.520 (0.060)$ & $1.090 (0.130)$ \\
\midrule
\multirow{3}{*}{\begin{tabular}[c]{@{}c@{}}\textit{In-Person} \\ \textit{User Study}\end{tabular}}
& BAIT-trust  & $0.32 (0.17)$ & $0.02 (0.02)$ & $0.01 (0.01)$ & $1.00 (0.00)$ & $0.046 (0.006)$ & $0.584 (0.136)$ & $1.301 (0.325)$ \\
& BAIT-infl  & $0.72 (0.18)$ & $0.05 (0.02)$ & $0.02 (0.01)$ & $0.99 (0.01)$ & $0.046 (0.005)$ & $0.559 (0.118)$ & $1.241 (0.283)$ \\
& BAIT-adapt & $0.52 (0.26)$ & $0.04 (0.02)$ & $0.02 (0.01)$ & $0.99 (0.01)$ & $0.045 (0.005)$ & $0.554 (0.116)$ & $1.230 (0.278)$ \\
\bottomrule
\multicolumn{9}{p{0.85\textwidth}}{\footnotesize \textit{Note:} Results show posterior mean (std. dev.) evaluated step-wise across $600$ interactions ($30$ subjects${\times}20$ consecutive episodes). BCE/MSE: binary cross-entropy/mean-squared error; Cos.: cosine similarity; ADE/FDE: average/final displacement error at the 1- and
H-step horizon.} \\
\end{tabular}
\vspace{-0.3cm}    
\end{table*}

\begin{center}
\textbf{Hypothesis 1: Influence and Task Success.} 
\emph{In repeated encounters, static and trust policies are easily exploitable, 
whereas influence policies maintain influence. Adaptive arbitration dynamically shifts between these strategies.}
\end{center}

\subsection{Simulator and Adaptive Human Model} 
We instantiate the MOMDP as a two-lane merging task in a 2D driving simulator based on  CARLO~\cite{cao2020reinforcement} across $N=20$ consecutive episodes and conduct $30$ independent trials using random seeds. Each episode consists of $75$ steps with a step size of $\Delta t{=}1$ s. While the physical state resets each episode, the human's latent state $(\phi^{GT},z^{GT})$ persists to model continuous adaptation across interactions. These ground-truth parameters remain inaccessible to all controllers. 

Following recent findings that humans aggressively exploit predictable policies but become conservative under uncertainty~\cite{sagheb2025unified, sagheb2023towards}, our generative human model simulates more complex, non-linear dynamics than the robot's internal filter. Specifically, the simulated human updates its ground-truth belief $\phi^{GT}$ using the same kinematic moving-average formulation defined for the robot's geometric proxy ($b_y^t$ in~\cref{eq:belief_proxy}). Furthermore, while the robot's filter models the strategy attractor ($z_{\text{tgt}}$ in~\cref{eq:z_ou}) as a simple interpolation of $\phi$, the actual human strategy $z^{GT}$ is drawn toward a highly non-linear ground-truth attractor $z_{\text{tgt}}^{GT}$.
To rigorously test our filter's tracking capability, $z_{\text{tgt}}^{GT}$ augments the baseline interpolation with two adversarial, rule-based overlays: a \emph{cautious bias} that smoothly shifts the human toward a conservative prototype when belief entropy is high ($\mathcal{H}(\phi^{GT}){>}0.75$), and an \emph{aggressive spike} that instantly snaps to an aggressive prototype if low entropy ($\mathcal{H}(\phi^{GT}){<}0.70$) persists over a moving window of $6$ steps. Without access to these underlying non-linear reactive thresholds, the robot's particle filter must robustly infer these rapid, strategic shifts using its simplified surrogate dynamics.

\subsection{Implementation Details.}
We instantiate the full BAIT pipeline using $K{=}100$ rollout sequences, $N_\phi{=}20$ long-term particles, and $N_z{=}50$ short-term particles. Specifically, each long-term particle tracks a single belief dimension $\phi{\in} [0,1]$, while each short-term particle tracks the 2-dimensional short-term strategy trade-off between forward progress ($z_{\text{progress}}$) and collision avoidance ($z_{\text{safety}}$) discussed in~\cref{sec:ProblemStatement}. These two adapted dimensions operate within 5-dimensional instantaneous cost vector ($d_z{=}5$) that also accounts for lane center deviation, heading, and steering smoothness. For the hierarchical particle filter, we maintain a history window $W{=}5$ with an exponential decay $\rho{=}0.7$. The short-term strategy employs an update rate $\alpha_z{=}0.8$, a mean-reverting rate $\alpha_{ou}{=}0.5$, and an Ornstein-Uhlenbeck noise scale $\sigma_z{=}0.1$. The long-term belief utilizes a geometric smoothing rate $\alpha_\phi{=}0.25$ and a scaling temperature $T_\phi{=}3.0$. The beta kernel is configured with a concentration $\kappa{=}50.0$, a slow reset probability $\epsilon{=}0.02$, and a uniform reset shape $c{=}1.0$. The critical interaction zone is triggered at a spatial threshold $d_{\text{interact}}{=}25.0$m. For the MPPI planner, the influence cost uses an entropy weight $w_e{=}0.8$ and directional weight $w_d{=}0.2$ while trust cost shifts these to $w_e{=}0.9$ and $w_d{=}0.1$. The CVaR safety constraint utilizes a risk-aversion parameter $\eta_r{=}1.0$, a per-step acceptable cost threshold $\tau{=}2.5$, and an effective infinite scalar penalty $M{=}10^9$.

\begin{figure*}[t]
    \centering
    \includegraphics[width=0.9\linewidth, trim=0 2mm 0 2mm, clip]{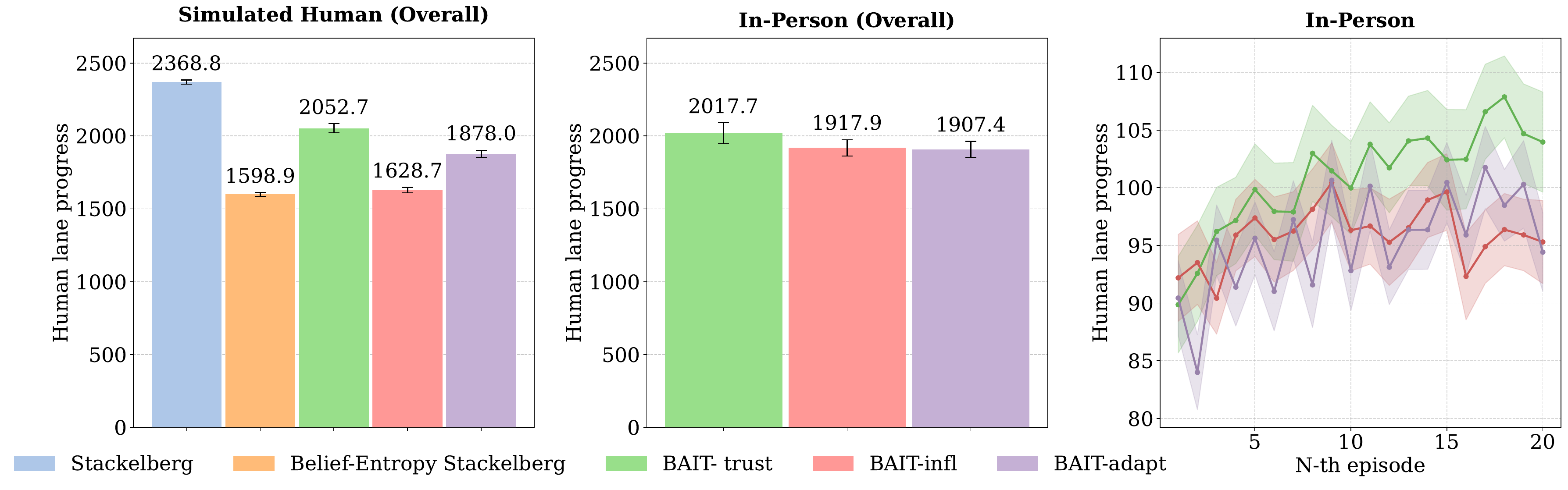}
    \vspace{-4mm}
    \caption{\textbf{Comparison of Human Lane Progress.} Overall lane progress for the BAIT variants (BAIT-trust, BAIT-infl, and BAIT-adapt) against Stackelberg baselines (Stackelberg, Belief-Entropy Stackelberg) when interacting with a belief-aware simulated human (left), alongside the overall (center) and episode-by-episode (right) progress of in-person users interacting with the three BAIT variants. Both the simulated and in-person data consists of $600$ total interactions ($30$ simulated/human subjects${\times} 20$ consecutive episodes). Error bars denote standard error. In simulation (left), both Belief-Entropy Stackelberg and BAIT-infl maintain low human lane progress by successfully influencing the human to yield. In contrast, the standard Stackelberg and BAIT-trust modes resulted in high human lane progress, as the human becomes assertive upon perceiving that the robot will yield. The BAIT-adapt mode balances these two extremes by strategically switching between influence and trust based on the recent interaction history. Furthermore, the in-person results (center, right) confirm that this strategic switching allows BAIT-adapt to consistently regulate human lane progress across repeated physical interactions.}
    \label{fig:sim_task_perf}
    \vspace{-0.5cm}    
\end{figure*}

\subsection{Baselines.}
We evaluate our BAIT framework against two baselines: \textbf{Stackelberg}, which assumes a stationary best-response human; and  \textbf{Belief-Entropy Stackelberg}, which augments the \textbf{Stackelberg} objective with a belief-entropy reward (similar to our influence cost in~\cref{eq:cost_infl}) to optimize strictly for latent influence. To ensure fair comparison, all controllers share the exact planning horizon and resolution ($H{=}5$, $\Delta t{=}1$), physical parameters, and reward configurations as BAIT. We also test two ablations to evaluate the extremes of our cost formulation: pure trust \textbf{BAIT-trust} ($\beta{=}0$) and pure influence \textbf{BAIT-infl} ($\beta{=}1$). Finally, \textbf{BAIT-adapt} arbitrates between episodes via binary, history-dependent switching ($\beta{\in}\{0,1\}$), shifting to influence mode ($\beta{=}1$) if the robot's success rate falls below $60\%$ over a sliding $5$-episode window or the human's belief indicates expected yielding ($\hat{\phi}{>}0.7$), and reverts to trust once both conditions clear. We evaluate the task and inference fidelity jointly to prevent the emergence of degenerate policies.

\subsection{Result and Discussions} 
\textbf{Estimation Robustness.} \cref{tab:pf_performance} demonstrates the policy-invariant tracking accuracy of BAIT's hierarchical particle filter. Because the Stackelberg baselines lack complete hierarchical latent state estimation, we compare the three BAIT variants to validate our filter robustly tracks human adaptation regardless of the robot policy, rather than serving as an ablation. Furthermore, presenting in-person tracking alongside simulated data validates our simulated human as a reliable proxy prior to discussion task outcomes.

Because ground-truth internal states are inaccessible for real human drivers, we extract offline pseudo-labels $(\hat{\phi}^{GT}, \hat{z}^{GT})$ over the recorded dataset by joint trajectory optimization. Because the belief $\phi$ only influences human actions indirectly, minimizing the negative log-likelihood (NLL) of actions is insufficient to uniquely recover the latent hierarchy. Thus, we extract the pseudo-labels by minimizing the action NLL subject to the structural priors of our generative hierarchy:
\begin{equation} \label{eq:pseudo_label}
\begin{split}
\min_{\hat{\phi}^{1:T}, \hat{z}^{1:T}} \sum_{t=1}^{T} \Big[ &-\log \pi_\mathcal{H}(a_\mathcal{H}^t \mid s^t, \hat{z}^t) \\
+ \lambda_z \|\hat{z}^t - z_{\text{tgt}}(\hat{\phi}^t)\|^2 
&+ \lambda_\phi (\hat{\phi}^t - b_y^t)^2 + \mathcal{C}_{\text{smooth}} \Big]
\end{split}
\end{equation}
where the first term fits the short-term strategy to the observed physical driving, the second term enforces the hierarchical coupling by anchoring $\hat{z}^t$ to its $\hat{\phi}^t$-conditioned prototype, the third term anchors the extracted belief to the geometric yield proxy $b_y^t$ (\cref{eq:belief_proxy}), and $\mathcal{C}_{\text{smooth}}$ applies standard temporal smoothing to prevent erratic, step-to-step fluctuations in both variables. Evaluating against these optimized pseudo-labels, the filter maintains tight convergence and low error across both environments, accurately capturing internal state transitions. Trajectory prediction remains flat across variants (H-ADE ${\approx }0.5$), confirming the expected posterior moments provide a reliable input to our deterministic surrogate regardless of $\beta$.

\textbf{Task performance.} To evaluate the robot's influence on human yielding, we report human lane progress analyzed via repeated-measures ANOVA ($F{=}250.51, p{<}0.001$) and Bonferroni-corrected post-hoc test ($\alpha{=}0.05$). Supporting \textbf{H1}, static \textbf{Stackelberg} and \textbf{BAIT-trust} allow significantly greater human lane progress than the influence-optimizing models, \textbf{Belief-Entropy Stackelberg} and \textbf{BAIT-infl}, (all pairwise $p{<}0.001$), as the human agent exploits their predictable yielding (\cref{fig:sim_task_perf}). Ablating the directional regularizer, a transparent-assertive variant ($w_e \mathcal{H}(\hat{\phi}^{k+1})^2+w_d (\hat{\phi}^{k+1})^2$) yields total lane progress $1962.64\pm36.58$, matching $\textbf{BAIT-trust}$ ($1957.66$), not \textbf{BAIT-infl} ($1533.74$), so predictability, not belief direction, governs influence. Furthermore, \textbf{BAIT-adapt} successfully arbitrates between influence and trust, achieving a statistically significant reduction in lane progress compared to \textbf{Stackelberg} and $\textbf{BAIT-trust}$ (both $p{<}0.001$) yet significantly higher than \textbf{BAIT-infl} and \textbf{Belief-Entropy Stackelberg} (both $p{<}0.001$). Crucially, BAIT achieves this effective behavioral arbitration without sacrificing real-time scalability: by propagating a noise-free deterministic surrogate rather than branching full trees like online MOMDP solvers (e.g., POMCPOW) or nested best-response MPCs, BAIT bypasses continuous-space branching bottlenecks to achieve real-time execution at ${\approx} 0.11$~s/step.

\section{User Study}\label{sec:user_study}

\begin{figure*}[t]
    \centering
    \includegraphics[width=\linewidth]{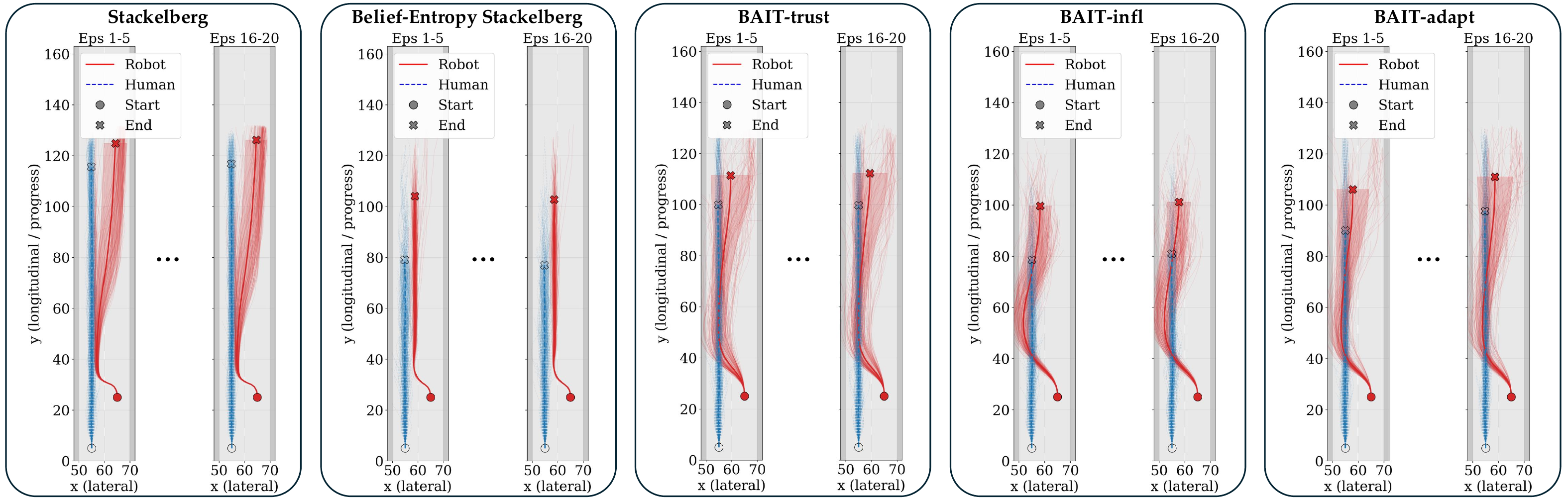}
    \vspace{-0.5cm}
    \caption{\textbf{Evolution of Interaction trajectories.} Comparison of robot (solid \textcolor{red}{red}) and simulated human (dashed \textcolor{blue}{blue}) trajectories during early (Eps $1-5$) and late (Eps $16-20$) interactions. Results are aggregated over $30$ independent trials across five control modes. While standard Stackelberg and BAIT-trust initially secure the lane merge, they fail to prevent the human from becoming aggressive over repeated interactions. Conversely, the Belief-Entropy Stackelberg, BAIT-infl, and BAIT-adapt controllers successfully maintain the human's yielding behavior across the majority of trials.} 
    \label{fig:bait_traj}
    \vspace{-0.5cm}    
\end{figure*}

We conducted an in-person user study to assess subjective trust and perceived comfort across three BAIT variants (\textbf{BAIT-trust}, \textbf{BAIT-infl}, and \textbf{BAIT-adapt}). Because our simulation result in~\cref{sec:simulation} confirmed equivalent efficacy between \textbf{Belief-Entropy Stackelberg} and \textbf{BAIT-infl} ($p=1.000$), we focus our experimental evaluation on three BAIT controllers. Participants interacted with the merging robot within the environment described in~\cref{sec:simulation}, controlling the human vehicle using a Logitech G$29$ steering wheel and pedal set (\cref{fig:bait_overview}). To ensure consistency, all implementation parameters remained identical to those used in the simulation experiments.

\vspace{-1mm}
\begin{center}
\textbf{Hypothesis 2: Human Trust and Comfort.} 
\emph{Transparency builds human trust and comfort, whereas unpredictability degrades them. Adaptive arbitration exerts necessary influence without sacrificing these subjective metrics.}
\end{center}
\vspace{-10pt}

\subsection{Participants}
We recruited $30$ participants ($11$ female, age $28{\pm} 6.2$), $14$ of whom reported prior experience riding in or driving alongside autonomous vehicles. This study was conducted in strict accordance with Institutional Review Board guidelines (IRB \#$25$-$0685$). Following a within-subject design, each participant completed $20$ consecutive merging episodes against each of the three controllers. Participants rated four custom-designed measures on a $7$-point Likert scale: comfort (``I felt comfortable driving alongside this autonomous vehicle (AV)"), discomfort (``Interacting with this AV made me feel tense or uneasy," reverse-coded as a consistency check), trust (``I trusted this AV to behave safely during our interactions"), and anticipation (``I could anticipate what AV was going to do next"). The presentation order was fully counter-balanced to mitigate ordering effects.

\subsection{Results and Discussions}


We report subjective outcomes evaluated using non-parametric Friedman tests and Bonferroni-corrected Wilcoxon post-hoc analyses ($\alpha{=}0.0167$). The in-person experimental results closely mirror our simulation results (\cref{fig:sim_task_perf}). \textbf{BAIT-trust} allowed the highest human lane progress, which increased over time as participants exploited the robot's transparent yielding. Conversely, both \textbf{BAIT-infl} and \textbf{BAIT-adapt} successfully asserted merging to maintain lower human lane progress over the episodes. Notably, \textbf{BAIT-adapt} achieved a statistically significant reduction compared to \textbf{BAIT-trust} ($p{<}0.01$), while demonstrating equivalent efficacy to \textbf{BAIT-infl} ($p{=}0.811$).

The user study results comparing the human trust,  comfort, and anticipation for the three controller variants also supported \textbf{H2}. The participants felt significantly more comfortable with the predictable yielding of \textbf{BAIT-trust} (comfort median $5$, discomfort median $3$) compared to \textbf{BAIT-infl} (comfort median $2$, discomfort median $5$) or \textbf{BAIT-adapt} (comfort median $4$, discomfort median $5$) significantly degrading overall comfort and discomfort ($p{<}0.005$ across all four pairwise comparisons). However, the trust variance was highly significant ($\chi^2(2){=}14.04,p{<}0.001$). While \textbf{BAIT-trust} maximized trust (median $5$), \textbf{BAIT-adapt} (median $3$) preserved user trust at level significantly higher than the pure influence baseline \textbf{BAIT-infl} (median $2$, $p{=0}.0128$). 
Importantly, this trust preservation coexists with effective influence maintenance. Because \textbf{BAIT-adapt} switches between modes across episodes via hysteresis, participants perceived it as less predictable (anticipation median $4$ vs. $6$ for \textbf{BAIT-trust}, $p{=0}.003$). These results reveal that while controlled adaptation must inherently sacrifice the human comfort to maintain long-term influence for task efficiency adversarial exploitation, allowing \textbf{BAIT-adapt} to make the human yield as effectively, it successfully mitigates degradation of human trust caused by pure influence.

\section{Real-World Experiment}\label{sec:realworld_exp}
To validate real-time deployment, we implemented the proposed \textbf{BAIT-adapt} controller on a physical Polaris GEM e2. The ego vehicle interacted with a human-driven GEM e4 in a scaled lane-merging scenario (Similar to Section~\ref{sec:simulation} and Section~\ref{sec:user_study}). State estimation was achieved using a Septentrio AsteRx SBi3 Pro+ GNSS receiver with a ProPak 6 RTK base station, eliminating localization uncertainty while low-level steering and throttle commands were executed via ROS PACMod. As shown in \cref{fig:GEM}, the \textbf{BAIT-adapt} runs in real time onboard the ego vehicle and successfully secures the merge. The robot smoothly asserts its right-of-way and consistently prompts the human driver to yield across repeated interactions, demonstrating the practical viability of the BAIT.

\section{Conclusion}
We introduce BAIT, a real-time controller for repeated human-robot interactions that integrates a hierarchical particle filter--tracking the human's short-term strategy and long-term perception--with a deterministic surrogate MPPI planner. BAIT explicitly arbitrates between long-horizon influence and human trust, while enforcing immediate task performance as a strict CVaR constraint. Across simulations, a user study, and real-world GEM vehicle deployments, BAIT maintains influence by consistently prompting humans to yield while mitigating the severe trust degradation of pure influence controllers. Although mode switching inherently sacrifices the human comfort of a trust policy, it achieves controlled unpredictability that preserves influence. Future work will replace the current binary-mode switching by optimizing the arbitration variable continuously to enable smoother adaptation.
\section*{Acknowledgments}
This work was supported by the National Science Foundation under Grant No. $2246448$ and the Center for Autonomy at the University of Illinois Urbana-Champaign.



\bibliographystyle{IEEEtran}
\bibliography{reference}

\end{document}